\documentclass{article}
\usepackage{graphicx}
\usepackage[tight]{subfigure}
\usepackage{color}
\usepackage{arxiv}
\usepackage[utf8]{inputenc} 
\usepackage[T1]{fontenc}    
\usepackage{hyperref}       
\usepackage{url}            
\usepackage{booktabs}       
\usepackage{amsfonts}       
\usepackage{nicefrac}       
\usepackage{microtype}      
\usepackage{bm}
\usepackage{lipsum}
\usepackage{amsmath}
\usepackage{paralist}
\usepackage{multirow}
\usepackage{comment}
\usepackage[ruled, vlined]{algorithm2e}
\usepackage{multicol}
\usepackage{textcomp, gensymb}
\newcommand{\argmax}{\mathop{\rm arg~max}\limits}

\newcommand{\expected}{\mathop{\mathbb{E}}\limits}

\title{An FPGA-Based On-Device Reinforcement Learning Approach using 
Online Sequential Learning}
\author{
  Hirohisa Watanabe \\
  Keio University\\
  3-14-1 Hiyoshi, Kohoku-ku, Yokohama, Japan\\
  \texttt{watanabe@arc.ics.keio.ac.jp}\\
  \And
  Mineto Tsukada\\
  Keio University\\
  3-14-1 Hiyoshi, Kohoku-ku, Yokohama, Japan\\
  \texttt{tsukada@arc.ics.keio.ac.jp}\\
  \And
  Hiroki Matsutani \\
  Keio University\\
  3-14-1 Hiyoshi, Kohoku-ku, Yokohama, Japan\\
  \texttt{matutani@arc.ics.keio.ac.jp} \\
}
\begin{document}
\maketitle

\begin{abstract}
	DQN (Deep Q-Network) is a method to perform Q-learning for
	reinforcement learning using deep neural networks.
	DQNs require a large buffer and batch processing for an experience
	replay and rely on a backpropagation based iterative optimization,
	making them difficult to be implemented on resource-limited edge
	devices.
	In this paper, we propose a lightweight on-device reinforcement
	learning approach for low-cost FPGA devices.
	It exploits a recently proposed neural-network based on-device
	learning approach that does not rely on the backpropagation method
	but uses OS-ELM (Online Sequential Extreme Learning Machine) based
	training algorithm.
	In addition, we propose a combination of L2 regularization and
	spectral normalization for the on-device reinforcement learning so
	that output values of the neural network can be fit into a certain
	range and the reinforcement learning becomes stable.
	The proposed reinforcement learning approach is designed for
	PYNQ-Z1 board as a low-cost FPGA platform.
	The evaluation results using OpenAI Gym demonstrate that the proposed
	algorithm and its FPGA implementation
	complete a CartPole-v0 task 29.77x and 89.40x faster than a
	conventional DQN-based approach when the number of hidden-layer nodes
	is 64.
\end{abstract}


\section{Introduction}\label{sec:intro}

%
%
Reinforcement learning differs from a typical deep learning in that
agents themselves explore their environment and learn appropriate
actions.
This means that it learns correct actions while creating a dataset.
In DQN (Deep Q-Network) \cite{dqn}, Q-learning for reinforcement
learning is replaced with deep neural networks so that it can acquire
a high generalization capability by the deep neural networks.
In this case, continuous input values can be used as inputs.
Also, to reduce a dependence on a sequence of input data, an
experience replay technique \cite{experience_replay}, in which past
experiences including states, actions, and rewards are recorded in a
buffer and then randomly picked up for training, is typically used for
DQNs.
%
%
However, such DQNs are costly for resource-limited edge devices and a
standalone execution on edge devices is not feasible, because 
they rely on a backpropagation based training algorithm that
iteratively optimizes their weight parameters and the convergence is
sometimes time-consuming.

%
%
In this paper, we propose a lightweight on-device reinforcement
learning approach for resource-limited FPGA devices.
It exploits a recently proposed neural-network based on-device
learning approach \cite{tukada_oselm} that does not rely on the
backpropagation methods but uses OS-ELM (Online Sequential Extreme
Learning Machine) based training algorithm \cite{os_elm}.
Computational cost for this training algorithm is quite low, because
its weight parameters are analytically solved in a one-shot manner
without the backpropagation based iterative optimization.
In theory, it has been demonstrated that it can satisfy the universal
approximation theorem \cite{universal_approximation_theorem} as in
deep learning.

%
%
However, since the training algorithm of OS-ELM assumes single
hidden-layer neural networks, their output values tend to be unstable
in some cases, e.g., when they are overfit to some specific inputs
and/or when unknown patterns are fed.
In the case of reinforcement learning, one of crucial issues is that
an action acquisition with Q-learning becomes unstable.
To address this issue, this paper proposes a combination of L2
regularization and spectral normalization \cite{spectral_normalization} 
so that output values of the proposed OS-ELM Q-Network can be fit
into a certain range and the reinforcement learning becomes stable.
This enables us to implement the reinforcement learning on small-sized
FPGA devices for standalone execution on resource-limited edge
devices.
In this paper, the proposed reinforcement learning approach is
designed for PYNQ-Z1 board.
The evaluation results using OpenAI Gym show that the proposed
algorithm and its FPGA implementation 
complete a CartPole task 29.77x and 89.40x faster than a conventional
DQN when the number of hidden-layer nodes is 64.

%
%
The rest of this paper is organized as follows.
Section \ref{sec:prelim} introduces basic technologies behind our
proposal.
Section \ref{sec:propose} proposes the lightweight on-device
reinforcement learning approach and illustrates an FPGA implementation.
In Section \ref{sec:eval}, it is evaluated 
in terms of training curve and execution time to complete a CartPole task.
Section \ref{sec:conc} summarizes this paper.

\section{Preliminaries}\label{sec:prelim}

This section introduces (1) ELM (Extreme Learning Machine), (2)
OS-ELM (Online Sequential ELM), (3) ReOS-ELM (Regularized OS-ELM), and
(4) DQN (Deep Q-Network).

\subsection{ELM}\label{ssec:elm}


%
%
ELM \cite{elm} is a batch training algorithm for single hidden-layer
neural networks.
In this case, the network consists of input layer, hidden layer, and 
output layer (see Figure \ref{fig:elm} in a few pages later).
The numbers of their nodes are $n$, $\tilde{N}$, and $m$, respectively.

%
%
Assuming an $n$-dimensional input chunk
$\bm{x} \in \mathbb{R}^{k \times n}$ with batch size $k$ is given,
an $m$-dimensional output chunk $\bm{y} \in \mathbb{R}^{k \times m}$
is computed as follows.
\begin{equation}\label{eq:elm_predict}
	\bm{y} = G(\bm{x} \cdot \bm{\alpha} + \bm{b})\bm{\beta},
\end{equation}
where $G$ is an activation function,
$\bm{\alpha}\in\mathbb{R}^{n \times \tilde{N}}$ is an input weight
matrix between input and hidden layers,
$\bm{\beta}\in\mathbb{R}^{\tilde{N} \times m}$ is an output weight
matrix between hidden and output layers, and
$\bm{b}\in\mathbb{R}^{\tilde{N}}$ is a bias vector of the hidden
layer.

%
%
Assuming this neural network approximates an $m$-dimensional target
chunk (i.e., teacher data) $\bm{t} \in \mathbb{R}^{k \times m}$ with
zero error, the following equation is satisfied.
\begin{equation}\label{eq:elm_1}
	G(\bm{x}\cdot \bm{\alpha} + \bm{b})\bm{\beta} = \bm{t}
\end{equation}
Here, the hidden layer matrix is defined as 
$\bm{H} \equiv G(\bm{x} \cdot \bm{\alpha} + \bm{b})$.
The optimal output weight matrix $\hat{\bm{\beta}}$ is computed as
follows.
\begin{equation}\label{eq:elm_2}
	\hat{\bm{\beta}} = \bm{H}^{\dagger}\bm{t},
\end{equation}
where $\bm{H}^{\dagger}$ is a pseudo inverse matrix of $\bm{H}$,
which can be computed with matrix decomposition algorithms, such as
SVD and QRD (QR Decomposition).

%
%
In ELM algorithm, the input weight matrix $\bm{\alpha}$ is initialized
with random values and not changed thereafter.
The optimization is thus performed only for the output weight matrix
$\bm{\beta}$; thus, it is quite simple compared with backpropagation
based neural networks that optimize both $\bm{\alpha}$ and $\bm{\beta}$.
In addition, the training algorithm of ELM is not iterative; it
analytically computes the optimal weight matrix $\bm{\beta}$ for a
given input chunk in one shot, as shown in Equation \ref{eq:elm_2}.
That is, it can always obtain the optimal $\bm{\beta}$ in one shot,
unlike a typical gradient descent method that iteratively tunes the
parameters toward the optimal solution.

%
%
Please note that ELM is a batch training algorithm and it becomes
costly when the training data size grows sequentially.
This means that, when a new training data arrives, the whole dataset
including the new data must be retrained to update the model.
This issue is a limiting factor for reinforcement learning, which can
be addressed by OS-ELM.

\subsection{OS-ELM}\label{ssec:os_elm}

%
%
OS-ELM \cite{os_elm} is an online sequential version of ELM, which can
update the model sequentially using an arbitrary batch size.
Assuming that the $i$-th training chunk 
$\{\bm{x}_i \in \mathbb{R}^{k_i \times n}, \bm{t}_i \in \mathbb{R}^{k_i \times m}\}$
with batch size $k_i$ is given, we need to compute an output weight
matrix $\bm{\beta}_i$ that can minimize the following error.
\begin{equation}\label{eq:os_elm_error}
	\left(\begin{bmatrix}\bm{H}_0 \\ \vdots \\ \bm{H}_i\end{bmatrix} \bm{\beta}_i - \begin{bmatrix} \bm{t}_0 \\ \vdots \\ \bm{t}_i \end{bmatrix} \right)^{2},
\end{equation}
where $\bm{H}_i$ is defined as 
$\bm{H}_i \equiv G(\bm{x}_i \cdot \bm{\alpha} + \bm{b})$.

%
%
Assuming $\bm{P_i} \equiv \left({\begin{bmatrix} \bm{H_0} \\ \vdots \\ \bm{H_i}\end{bmatrix}}^{\top}\begin{bmatrix} \bm{H_0} \\ \vdots \\ \bm{H_i}\end{bmatrix}\right)^{-1} (i \ge 0)$,
the optimal output weight matrix is computed as follows.
\begin{equation}\label{eq:os_elm_0}
	\begin{split}
		\bm{P}_i &= \bm{P}_{i-1} - \bm{P}_{i-1}\bm{H}_i^{\top}\left(\bm{I} + \bm{H}_i\bm{P}_{i-1}\bm{H}_i^{\top}\right)^{-1}\bm{H}_i\bm{P}_{i-1}\\
		\bm{\beta}_i &= \bm{\beta}_{i-1} + \bm{P}_i\bm{H}_i^{\top}\left(\bm{t}_i - \bm{H}_i\bm{\beta}_{i-1}\right)
	\end{split}
\end{equation}



In particular, the initial values $\bm{P}_0$ and $\bm{\beta}_0$ are
precomputed as follows. This computation is called initial training.
\begin{equation}\label{eq:os_elm_1}
	\begin{aligned}
		\bm{P}_{0}     & = \left(\bm{H}_{0}^{\top}\bm{H}_{0}\right)^{-1} \\
		\bm{\beta}_{0} & = \bm{P}_{0}\bm{H}_{0}^{\top}\bm{t}_{0}         \\
	\end{aligned}
\end{equation}

As shown in Equation \ref{eq:os_elm_0}, the output weight matrix
$\bm{\beta}_i$ and its intermediate result $\bm{P}_i$ are computed
from the previous training results $\bm{\beta}_{i-1}$ and $\bm{P}_{i-1}$.
Thus, OS-ELM can sequentially update the model with a newly-arrived
target chunk in one shot, and
there is no need to retrain all the past data unlike ELM.

%
%
In this approach, the major bottleneck is the pseudo inverse operation
${\left(\bm{I} + \bm{H_{i}}\bm{P_{i-1}}\bm{H_{i}^{\top}}\right)}^{-1}$
in Equation \ref{eq:os_elm_0}.
As proposed in \cite{tukada_oselm}, the batch size $k$ is fixed at 1
in this paper so that the pseudo inverse operation of $k \times k$
matrix for the sequential training is replaced with a simple
reciprocal operation; thus, we can eliminate SVD or QRD
computation from Equation \ref{eq:os_elm_0}.

\subsection{ReOS-ELM}\label{ssec:r_oselm}

ReOS-ELM \cite{r_oselm} is an OS-ELM variant where an L2 regularization
is applied to the output weight matrix $\bm{\beta}$ so that it can
mitigate an overfitting issue of OS-ELM and improve its generalization
capability.
The training algorithm of ReOS-ELM is same as that of OS-ELM, except
that the initial training of $\bm{P}_0$ and $\bm{\beta}_0$ is changed
as follows.
\begin{equation}\label{eq:r_oselm_1}
	\begin{aligned}
		\bm{P}_{0}     & = \left(\bm{H}_{0}^{\top}\bm{H}_{0} + \delta \bm{I}\right)^{-1} \\
		\bm{\beta}_{0} & = \bm{P}_{0}\bm{H}_{0}^{\top}\bm{t}_{0},                        
	\end{aligned}
\end{equation}
where $\delta$ is a regularization parameter that controls an
importance of the regularization term.

\subsection{Reinforcement Learning and DQN}\label{ssec:dqn}

%
%
In DQNs, deep neural networks are used for Q-learning which is a
typical reinforcement learning algorithm.
In time step $t$, $Q_{\theta_1}(s_t, a_t)$ represents a value for
taking action $a_t$ in state $s_t$, predicted with a set of neural
network parameters $\theta_1$.
In this case, $\theta_1$ is trained so that the value 
$Q_{\theta_1}(s_t, a_t)$ can be predicted accurately by the neural network.
%
%
However, if $\theta_1$ is trained for each time step $t$, it is
continuously changed and the Q-learning will not be stable.
To address this issue, DQNs use a fixed target Q-network technique
\cite{dqn_nature}, in which another neural network with a set of
parameters $\theta_2$ is used for stabilizing the Q-learning, 
in addition to that with $\theta_1$.
More specifically, $\theta_2$ is used but fixed for a while, and it is
updated with $\theta_1$ at a predefined interval.

In DQNs, an optimization target is computed as follows.
\begin{equation}\label{eq:dqn_target}
	f(r_t, s_{t+1}, d_t) = r_t + (1-d_t)\gamma \max_{a\in A}{Q_{\theta_2}\left(s_{t+1}, a\right)},
\end{equation}
where $\gamma \in [0,1]$ is a discount rate that controls an
importance of the next step, $r_t$ is a current reward given by an
environment, and $d_t$ indicates if the current episode \footnote{In
	this paper, an episode is defined as a complete sequence of states,
actions, and rewards.} is finished or not.
If $d_t$ is equal to 1, the current episode is finished and a new
episode is started.
As shown in Equation \ref{eq:dqn_target}, the sum of the reward and
the maximum Q-value among all the possible actions $A$ in one step
ahead is regarded as the optimization target.
As mentioned above, $\theta_2$ is periodically updated with $\theta_1$ 
by using the fixed target Q-network technique.
%
%
Specifically, the loss value for $\theta_1$ is denoted as follows
\cite{spinningup}.
\begin{equation}\label{eq:dqn_1}
	L(\theta_1) = \expected_{(s_t, a_t, r_t, s_{t+1}, d_t)\sim D}{\left[\biggl(Q_{\theta_1}(s_t, a_t) - f(r_t, s_{t+1}, d_t)\biggr)^2 \right]},
\end{equation}
where $D$ is a buffer for the experience replay technique \cite{dqn},
which is used to suppress impacts of temporal dependence on input data
for training.
In this case, past experiences (e.g., $s_t$, $a_t$, $r_t$,
$s_{t+1}$, and $d_t$ in Equation \ref{eq:dqn_1}) are stored in the
buffer $D$.
Then, they are randomly picked up from the buffer to form a batch
which will be used for updating the weight parameters of the neural
network.
							
\subsection{Spectral Regularization and Spectral Normalization}
\label{ssec:spectral_normalization}
							
%
%
To stabilize an action acquisition with Q-learning, we focus on
regularization methods used in deep learning.
Specifically, for reinforcement learning, a range of neural network outputs should be within a
constant multiplication of their input for the stability.
Such a property is referred to as Lipschitz continuity.
More specifically, assuming an input value is changed from $x_1$ to
$x_2$, their output values $f(x_1)$ and $f(x_2)$ should satisfy the
following constraint.
\begin{equation}\label{eq:spectral_1}
	\forall x_1, x_2, \, \|f(x_1) - f(x_2)\| \leq K\|x_1 - x_2\|,
\end{equation}
where $K \in \mathbb{R}$ is a constant value called Lipschitz constant.
Lipschitz constant of a neural network is derived by partial products
of Lipschitz constants of all the layers,
each of which is equal to a product of Lipschitz constant of a weight
matrix (i.e., its largest singular value) and that of an activation
function (i.e., $\leq 1$ for ReLU and tanh).
It should be suppressed for the stable Q-learning.
A spectral regularization \cite{spectral_regularization} can be used
to suppress the Lipschitz constant of a neural network, in which the
sum of the largest singular value in each weight matrix is added to
the loss function as a penalty term.
							
%
%
In practice, a well-known extension of the spectral regularization is
spectral normalization \cite{spectral_normalization}, in which an
output of a neural network is computed based on partial products of
input data and each weight matrix divided by its largest singular
value.
In this case, the Lipschitz constant is limited to $\leq 1$.
Since 1-Lipschitz continuity is required for GANs (Generative
Adversarial Networks), it is widely used in these applications.
In this paper, we use this approach for stabilizing the OS-ELM based
reinforcement learning.

\section{On-Device Reinforcement Learning Approach}\label{sec:propose}

In Q-learning, the value $Q_{\theta}(s_t, a_t)$ is approximated with a
neural network.
Toward the standalone reinforcement learning on resource-limited edge
devices, in this paper we propose to use OS-ELM for this purpose.

\subsection{Baseline OS-ELM Q-Network}\label{ssec:oselm_q_algorithms}

\begin{figure*}[tb!]
	\centering
	\begin{minipage}{0.295\hsize}
		\centering
		\includegraphics[height=50mm]{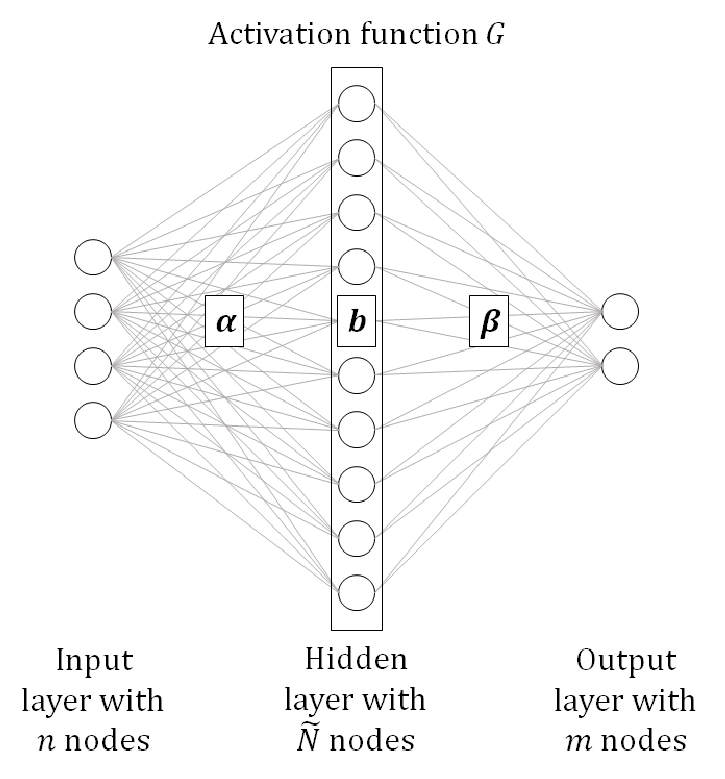}
		\caption{Extreme Learning Machine}
		\label{fig:elm}
	\end{minipage}
	\begin{minipage}{0.7\hsize}
		\includegraphics[height=50mm]{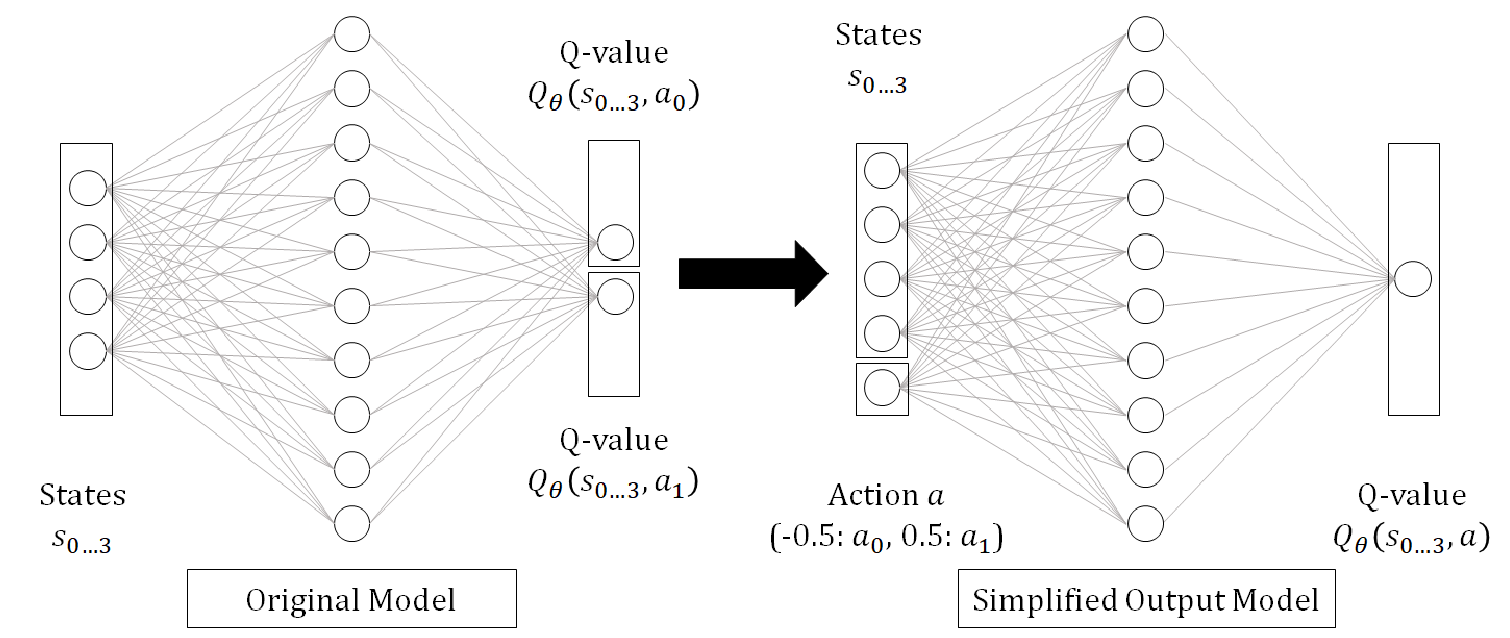}
		\caption{Simplified output model (numbers of state variables
		and actions are 4 and 2 in this example)}
		\label{fig:Q_model_change}
	\end{minipage}
\end{figure*}

\begin{algorithm}[t!]
	\caption{OS-ELM Q-Network} 
	\label{alg:oselm_q_network}
	\DontPrintSemicolon
	\nl Initialize parameters $\theta_1 = \{\bm{\alpha_0}, \bm{\beta_0}\}$ using random values $\mathbb{R} \in [0,1]$ \\
	\nl $\sigma_{max}(\bm{\alpha_0}) \leftarrow \text{SVD} (\bm{\alpha_0})$ \\
	\nl $\bm{\alpha_0} \leftarrow \bm{\alpha_0} / \sigma_{max}(\bm\alpha_0)$ \tcp{Initialize $\bm{\alpha_0}$}
	\nl Initialize parameters $\theta_2$ as $\theta_2 \leftarrow \theta_1$ \\
	\nl Initialize buffer $D$ \\
	\nl Initialize global step $t$ \\
	\nl \For{$episode \in 1\dots$}{
		\nl \For{$step \in 1\dots$}{
			\nl $t \leftarrow t + 1$ \\
			\tcp{Determine}
			\nl \If{random value $r_1 < \varepsilon_1$}{
				\nl$a_t \leftarrow \argmax_{a\in A}{Q_{\theta_1}(s_t, a)}$ \\
			}
			\nl \Else{
				\nl$a_t \leftarrow $ random action value\\
			}
			\tcp{Observe}
			\nl Observe $(s_{t+1}, r_t, d_t)$ from environment\\
			\nl \If{$d_t == 1$}{
				\nl Break
			}
			\tcp{Store}
			\nl Store $(s_t, a_t, r_t, s_{t+1}, d_t)$ in buffer $D$ \\
			\tcp{Update}
			\nl \If{$t == \tilde{N}$}{
				\nl Retrieve $\forall i \in [1,\tilde{N}], (s_i, a_i, r_i, s_{i+1}, d_i)$ from buffer $D$ \\
				\nl Update   $\forall i \in [1,\tilde{N}], Q_{\theta_1}(s_i, a_i)$ to $clip(-1,$ $ r_i + (1-d_i) \gamma \max_{a\in A}{Q_{\theta_2}(s_{i+1}, a)},1)$ \tcp{Initialize $\bm{\beta_t}$}
			}
			\nl \ElseIf{$t > \tilde{N}$}{
				\nl\If{random value $r_2 < \varepsilon_2$}{
					\nl Update $Q_{\theta_1}(s_t, a_t)$ to $clip(-1,r_t + (1-d_t) \gamma \max_{a\in A}{Q_{\theta_2}(s_{t+1},a)},1)$ \tcp{Update $\bm{\beta_t}$}
				}
			}
		}
		\nl \If{$episode \, \% \, UPDATE\_STEP == 0$}{
			\nl$\theta_2 \leftarrow \theta_1$ \\
		}
	}
\end{algorithm}

%
%
Algorithm \ref{alg:oselm_q_network} shows the proposed OS-ELM
Q-Network.
It consists of four states: Determine, Observe, Store, and
Update.
\begin{itemize}
	\item In Determine state (lines 10-13), a current action $a_t$ is
	      determined based on the current state $s_t$.
	      More specifically, an action that maximizes the Q-value (line 11) or
	      randomly-selected one (line 13) is selected as $a_t$.
	\item In Observe state (lines 14-16), based on an interaction using
	      $a_t$ with the environment, the next state $s_{t+1}$, reward $r_t$,
	      and flag $d_t$ are observed.
	\item In Store state (line 17), these observed values, action $a_t$,
	      and state $s_t$ are stored in buffer $D$ so that they can be used in
	      Update state.
	\item In Update state (lines 18-23), $\bm{\beta}$ is initialized or
	      updated, depending on the global step $t$.
	      More specifically, it is initially trained with stored values in $D$
	      (line 20) based on Equation \ref{eq:os_elm_1} when the number of
	      experiences in $D$ is same as $\tilde{N}$ (i.e., $t == \tilde{N}$).
	      Or, it is sequentially updated with the latest experience (line 23)
	      based on Equation \ref{eq:os_elm_0} when $t > \tilde{N}$.
	      The former is referred as an initial training and the latter is
	      referred as a sequential training.
\end{itemize}
Please note that the buffer $D$ is used for the initial training only
and it is not used in subsequent sequential training in the case of
OS-ELM Q-Network.

%
%
\paragraph{Fixed Target Q-Network}
OS-ELM Q-Network uses the fixed target Q-network technique as well as
DQNs.
At first, two sets of neural network parameters $\theta_1$ and
$\theta_2$ are initialized in lines 1 and 4.
$\theta_1$ is updated more frequently (lines 20 and 23) and $\theta_2$
is synchronized with $\theta_1$ at a certain interval (lines 24-25).
Please note that a straightforward algorithm that approximates
$Q(s_t, a_t)$ with OS-ELM is unstable and cannot complete a
reinforcement learning task in this paper.
We thus introduce some techniques below in order to improve OS-ELM
Q-Network.

%
%
\paragraph{Simplified Output Model}\label{para:change_model_shape}
In DQNs, the $i$-th node of an output layer is Q-value of the $i$-th
action, and they are trained so that the $i$-th node can predict 
$Q(s, a_i)$ accurately.
In this case, their input and output sizes are equal to the numbers of
state variables and actions, respectively.
The left hand side of Figure \ref{fig:Q_model_change} shows an example
of such a network when the numbers of state variables and actions are
4 and 2, respectively.
Since an action value is fed to the model in this case, Q-value is 
calculated for all the possible actions.


In Update state of DQNs, a loss value computed with Equations
\ref{eq:dqn_target} and \ref{eq:dqn_1} is used for the backpropagation based iterative
optimization.
In OS-ELM, on the other hand, teacher data $\bm{t} \in \mathbb{R}^{m}$
is required to update $\bm{\beta}$ when the batch size $k$ is 1, as
shown in Equation \ref{eq:os_elm_0}.
To directly use
$\left(r_t + \left(1-d_t\right) \gamma \max_{a\in A}{Q_{\theta_2}\left(s_{t+1}, a\right)}\right)$
in Equation \ref{eq:dqn_target} to update $\bm{\beta}$, in this paper we
employ a simplified output model, which is illustrated in the right
hand side of Figure \ref{fig:Q_model_change}.
In this model, a set of state variables and an action value (e.g.,
-0.5 for action $a_0$ and 0.5 for action $a_1$) is given as an input
and its corresponding Q-value is an output, which is scalar (i.e., $m=1$).
Thus,
$\left(r_t + \left(1-d_t\right) \gamma \max_{a\in A}{Q_{\theta_2}\left(s_{t+1}, a\right)}\right)$
can be directly used as a teacher data when updating $\bm{\beta}$ in
the simplified output model (lines 20 and 23).


%
%
\paragraph{Q-Value Clipping}\label{para:q_value_clipping}
OS-ELM Q-Network tends to be unstable especially when unseen inputs
are fed to the network, and its output values become anomaly in such
cases.
Such outliers hinder the reinforcement learning, because these values
are significantly large and exceed a range of normal reward values.
In a typical setting for the reinforcement learning, the maximum
reward given by the environment is 1 and the minimum reward is -1.
%
%
Thus, as shown in lines 20 and 23, output values of OS-ELM Q-Network
are clipped so that they are fit into the range of
$-1 \leq r_t + (1 - d_t) \gamma \max_{a \in A}{Q_{\theta_2}(s_{t+1}, a)} \leq 1$.
Such a Q-value clipping suppresses outliers and enables a stable
reinforcement learning with OS-ELM Q-Network.

%
%
\paragraph{Random Update}\label{para:random_update}
DQNs typically train their neural network parameters in a batch manner
and use the experience replay technique to form a batch randomly so
that it can mitigate a dependence on a sequence of input data.
On the other hand, OS-ELM is a sequential training algorithm that can
update its neural network parameters sequentially with a small batch
size $k$.
As mentioned in Section \ref{ssec:os_elm}, the major bottleneck of
OS-ELM when implemented for resource-limited FPGA devices is the
pseudo inverse matrix operation that may require an SVD or QRD core.
In \cite{tukada_oselm}, the pseudo inverse matrix operation is
eliminated by fixing $k$ to 1 for enabling the neural network based
on-device learning.
In this paper, to reduce the dependence on a sequence of input data
while keeping the small batch size $k$ to 1, we adopt a method of
randomly determining whether or not to update the neural network
parameters for each step, as shown in lines 22-23.
More specifically, depending on a random value $r_2$, the latest
experience (i.e., a set of observed values, action $a_t$, and state
$s_t$) is sequentially trained so that the batch size is fixed to 1
and the pseudo inverse matrix operation can be eliminated.
Assuming that the first initial training is done by software and
all the subsequent sequential training is computed by the FPGA device
(see Figure \ref{fig:design_overview}), we can eliminate the buffer
$D$ in the FPGA part.
Thus, a combination of the random update with OS-ELM whose batch
size is set to 1 \cite{tukada_oselm} can reduce both computational
cost and memory usage \footnote{
This approach can mitigate temporal dependency, but the 
sampling efficiency is reduced compared to the experience replay.
}.

\subsection{OS-ELM Q-Network with Regularization/Normalization}
\label{ssec:spectral_normalization_with_l2}


%
%
In Q-learning, a neural network is updated based on comparisons of an
expected value of the reward with the next state; thus, it can be
expected that Q-values in successive states are basically close to
recent ones.
As mentioned in Section \ref{ssec:spectral_normalization}, the
spectral regularization and normalization would be effective in
reinforcement learning for improving the generalization capability.
As discussed below, our recommendation is that the spectral
normalization and the L2 regularization are applied to weight
parameters $\bm{\alpha}$ (lines 2-3) and $\bm{\beta}$ (line 20),
respectively.

%
%
\paragraph{Spectral Normalization for $\bm{\beta}$}
Let us start with the spectral normalization for the weight parameter
$\bm{\beta}$ of OS-ELM Q-Network.
Let $\sigma_{max}({\bm{\beta}}_i)$ is the largest singular value in
$\bm{\beta}$ at step $i$.
In this case, ${\bm{\beta}}_i$ is divided by $\sigma_{max}({\bm{\beta}}_i)$
for every feedforward operation.
To obtain $\sigma_{max}(\bm{\beta}_i)$, SVD is typically applied to
$\bm{\beta}$ for every time, which is a costly operation; so, we do
not use the spectral normalization for $\bm{\beta}$.

\paragraph{L2 Regularization for $\bm{\beta}$}
In this paper, we thus use the L2 regularization for $\bm{\beta}$ as
an alternative to the spectral normalization for $\bm{\beta}$.
In this case, the initial training of Equation \ref{eq:os_elm_1},
which is called from line 20 of Algorithm \ref{alg:oselm_q_network},
is replaced with Equation \ref{eq:r_oselm_1}.
This approach is validated below.
Assuming $\bm{A}$ is a general matrix, the following relation is
satisfied.
\begin{equation}\label{eq:spectral_2}
	\|\bm{A}\|_2^2 = \sigma^2_{max}(\bm{A}) \leq \|\bm{A}\|_F^2 = \sum_{i}{\sigma^2_{i}{(\bm{A})}},
\end{equation}
where $\|\cdot\|_2$ and $\|\cdot\|_F$ denote a spectral norm and an L2
norm, respectively.
As shown in Relation \ref{eq:spectral_2}, the L2 norm introduces a
stronger constraint than the spectral norm \cite{spectral_regularization}.
This means that the L2 regularization for $\bm{\beta}_i$ of OS-ELM can
introduce the same or stronger effect of the spectral regularization.
%
%
\paragraph{Spectral Normalization for $\bm{\alpha}$}
Different from $\bm{\beta}$, weight parameter $\bm{\alpha}$ of OS-ELM
is randomly generated at the initialization step and not changed at
runtime.
Since the initial values of $\bm{\alpha}$ can be computed at offline
(e.g., by software), the spectral normalization can be easily applied
to $\bm{\alpha}$, as shown in lines 2-3 of Algorithm \ref{alg:oselm_q_network}.
By applying the spectral normalization for $\bm{\alpha}$, the
Lipschitz constant depending on $\bm{\alpha}$ is suppressed within 1
or less; 
thus, the Lipschitz constant of OS-ELM is $\sigma_{max}(\bm{\beta}_i)$
or less.
More specifically, it depends on $\bm{\beta}_i$ and the L2
regularization parameter $\delta$, which means that the Lipschitz
constant can be controlled by these parameters.
As a result, by a combination of the spectral normalization for
$\bm{\alpha}$ and the L2 regularization for $\bm{\beta}$, the
Lipschitz constant of OS-ELM can be kept under $\sigma_{max}(\bm{\beta}_i)$.
\footnote{
	As mentioned in Section \ref{ssec:spectral_normalization}, 
	when ReLU or tanh is used as an activation function, 
	Lipschitz constant of a neural network is derived 
	as a partial product of Lipschitz constant of each layer. 
	In this case, Lipschitz constant of the original network 
	without regularization/normalization at step $i$ is derived as 
	$\sigma_{max}(\bm{\alpha})\sigma_{max}(\bm{\beta}_i)$.
}
\subsection{FPGA Implementation}\label{ssec:impl}

\begin{figure}[tb]
	\centering
	\includegraphics[height=60mm]{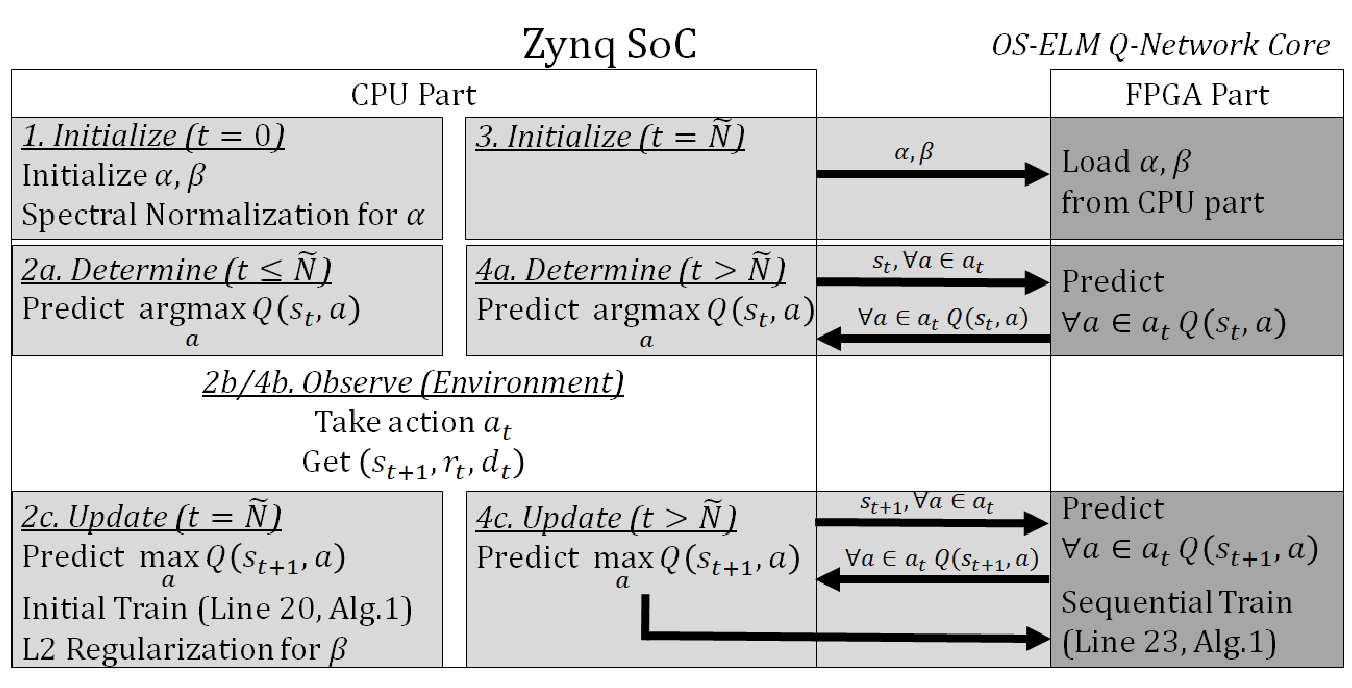}
	\caption{On-device reinforcement learning on PYNQ-Z1 platform
		(Steps 4a and 4c are OS-ELM Q-Network core implemented 
	in FPGA part)}
	\label{fig:design_overview}
\end{figure}

%
%
Table \ref{tb:env} shows the target platform in this paper.
Figure \ref{fig:design_overview} shows the design overview of 
{\bf FPGA} that consists of CPU and FPGA parts.
The predict and sequential train modules in Steps 4a and 4c are
designed with Xilinx Vivado and implemented in a programmable
logic part (denoted as FPGA part) of PYNQ-Z1 platform, while the
initial train in Step 2c is executed by the CPU part (i.e., Cortex-A9
processor).
After the initial train (Step 2c), Steps 4a, 4b, and 4c are
continuously executed as a main loop.
We assume that the interactions with environment (Steps 2b and 4b) are
handled by the CPU part.
	
%
%
A low-cost OS-ELM core optimized to batch size 1 was proposed in
\cite{tukada_oselm}.
In this paper, we redesigned a further optimized core that includes
both the predict and sequential train modules (i.e., Steps 4a and 4c)
in Verilog HDL,
and it is implemented for the same FPGA platform as in \cite{tukada_oselm}.
The target FPGA device is Xilinx XC7Z020-1CLG400C.
Operating frequency of the programmable logic part is 100MHz, while
the CPU is running at 650MHz.
Xilinx Vivado v2017.4 is used for the implementation.
		
%
%
As shown in the right hand side of Figure \ref{fig:Q_model_change}, in
the OS-ELM Q-Network core, its input size (i.e., the number of
input-layer nodes) is equal to the sum of the numbers of state
variables and a single action variable, which is five in the
CartPole-v0 task.
The output size is 1, which is a scalar.
The number of hidden-layer nodes is varied from 32 to 256 in the
evaluations.
The predict and sequential train modules can be implemented with
matrix add, mult, and div operations.
SVD or QRD core is not needed as in \cite{tukada_oselm}.
For these matrix operations, only a single set of add, mult, and div
units is implemented in this design for minimizing the area, but a
parallel execution using multiple arithmetic units is also possible.
We use 32-bit Q20 numbers as a fixed-point number format.
Input data, weight parameters $\bm{\alpha}$ and $\bm{\beta}$, and
intermediate computation results are stored in on-chip BRAMs.
As mentioned in Section \ref{ssec:oselm_q_algorithms}, since the fixed
target Q-network technique is used, two sets of neural network
parameters $\theta_1$ and $\theta_2$ are needed.
Specifically, the same $\bm{\alpha}$ is used for both $\theta_1$ and
$\theta_2$, while different $\bm{\beta}$ is needed for $\theta_1$ and
$\theta_2$; thus, two sets of $\bm{\beta}$ are implemented in the BRAMs.

\begin{table}[bt]
        \centering
        \caption{Specification of target platform}
        \label{tb:env}
        \begin{tabular}{c|c} \hline\hline
        OS   & PYNQ Linux based on Ubuntu 18.04 \\
        CPU  & Cortex-A9 processor (650MHz)     \\
        RAM  & DDR3 SDRAM (512MB)               \\
        FPGA & Zynq XC7Z020-1CLG400C (100MHz)   \\\hline
        \end{tabular}
\end{table}
\begin{table}[bt]
	\centering
	\caption{FPGA resource utilization of OS-ELM Q-Network core}
	\label{tb:fpra_resouce}
	\begin{tabular}{c|r|r|r|r}\hline\hline
		$\tilde{N}$ & BRAM [\%] & DSP [\%] & FF [\%] & LUT [\%] \\ \hline
		32          &  2.86     & 1.82     & 1.49    &  3.52   \\
		64          & 11.43     & 1.82     & 2.47    &  5.00   \\
		128         & 45.71     & 1.82     & 4.50    &  7.93   \\
		192         & 91.43     & 1.82     & 6.44    & 11.01   \\ \hline
	\end{tabular}
\end{table}
		
%
%
Table \ref{tb:fpra_resouce} shows FPGA resource utilization of the
OS-ELM Q-Network core that consists of the predict and sequential
train modules when the number of hidden-layer nodes $\tilde{N}$ is
changed from 32 to 256.
The largest design with 256 hidden-layer nodes cannot be implemented
for PYNQ-Z1 board due to an excessive BRAM requirement.
The other designs can be fit into the FPGA device.
The BRAM utilization is thus a limiting factor, and those of the other
resources are not high.


\section{Evaluations}\label{sec:eval}

%
%
The proposed OS-ELM Q-Network is evaluated in terms
of the execution time to complete a reinforcement learning task.
Its variants with and without the spectral normalization and L2
regularization techniques are compared to a typical DQN.

\subsection{Evaluation Environment}\label{ssec:eval_env}

%
%
As a reinforcement learning task in this experiment, we use OpenAI
Gym CartPole-v0 that tries to make an inverted pendulum stand longer.
As simulation parameters, Cart position, Cart velocity, Pole angle,
and Pole velocity at tip are set to -2.4 to 2.4, $-\infty$ to $\infty$,
-41.8\degree to 41.8\degree, and $-\infty$ to $\infty$, respectively.
The numbers of state variables and actions are 4 and 2, respectively.

		
%
%
The following designs are compared in terms of
(i) training curve and
(ii) average execution time to complete the reinforcement learning task.
The proposed FPGA design is evaluated in terms of FPGA resource utilization.
\begin{enumerate}
	\item {\bf OS-ELM}:
	      The proposed OS-ELM Q-Network with the fixed target Q-network,
	      simplified output model, Q-value clipping, and random update techniques
	      (i.e., the L2 regularization and spectral normalization are not included)
	\item {\bf OS-ELM-L2}:
	      The above {\bf OS-ELM} with the L2 regularization for $\bm{\beta}$
	\item {\bf OS-ELM-Lipschitz}:
	      The above {\bf OS-ELM} with the spectral normalization for $\bm{\alpha}$
	\item {\bf OS-ELM-L2-Lipschitz}:
	      The above {\bf OS-ELM} with the spectral normalization for $\bm{\alpha}$
	      and L2 regularization for $\bm{\beta}$
	\item {\bf DQN}:
	      A three-layer DQN with the fixed target Q-network and experience
	      replay
	\item {\bf ELM}:
	      The above {\bf DQN} replaced with ELM with the simplified output model
	      and Q-value clipping
	\item  {\bf FPGA}:
	      Same as {\bf OS-ELM-L2-Lipschitz} but its prediction and sequential
	      training parts are implemented in programmable logic using
	      fixed-point numbers as described in Section \ref{ssec:impl}
\end{enumerate}
		
%
%
We use ReLU as an activation function.
As reinforcement learning parameters, we use the following setting:
$\varepsilon_1=0.7$, $\varepsilon_2=0.5$, and $UPDATE\_STEP=2$.
As the L2 regularization parameter, $\delta$ is set to 1 and 0.5 for
{\bf OS-ELM-L2} and {\bf OS-ELM-L2-Lipschitz}, respectively.
In {\bf DQN}, $\varepsilon_2$ is not used, the buffer depth for the
experience replay is set to 10,000, the batch size is set to 32, 
Adam \cite{adam} is used as an optimizer, the learning rate is set to 0.01,
and Huber function \cite{huber} is used as a loss function.

\subsection{Training Curve}\label{ssec:eval_curve}
	
%
%
In this section, algorithm-level evaluations for the reinforcement
learning task are conducted.
Among the seven designs listed in Section \ref{ssec:eval_env},
{\bf ELM}, {\bf OS-ELM}, {\bf OS-ELM-L2}, {\bf OS-ELM-Lipschitz},
{\bf OS-ELM-L2-Lipschitz}, and {\bf DQN} are compared \footnote{
Here, {\bf OS-ELM-L2-Lipschitz} is corresponding to {\bf FPGA}.
Their difference is that {\bf FPGA} uses 32-bit Q20
fixed-point numbers, but the negative impact was not significant in this
experiment.
}.
They are executed as a software on a 650MHz Cortex-A9 processor of the
PYNQ-Z1 board.
NumPy version 1.17.2 and Pytorch version 1.3.0 are used for DQN and
the ELM/OS-ELM based approaches, respectively.
%
%
In the designs other than {\bf DQN}, because their dependence on
initial weight parameters are high, unpromising weight parameters are
reset  when a given condition is met.
Specifically, the ELM/OS-ELM based approaches are reset if they did
not complete the reinforcement learning task after 300 episodes
elapsed.
		
\begin{figure*}[tb]
	\centering
	\includegraphics[height=120mm]{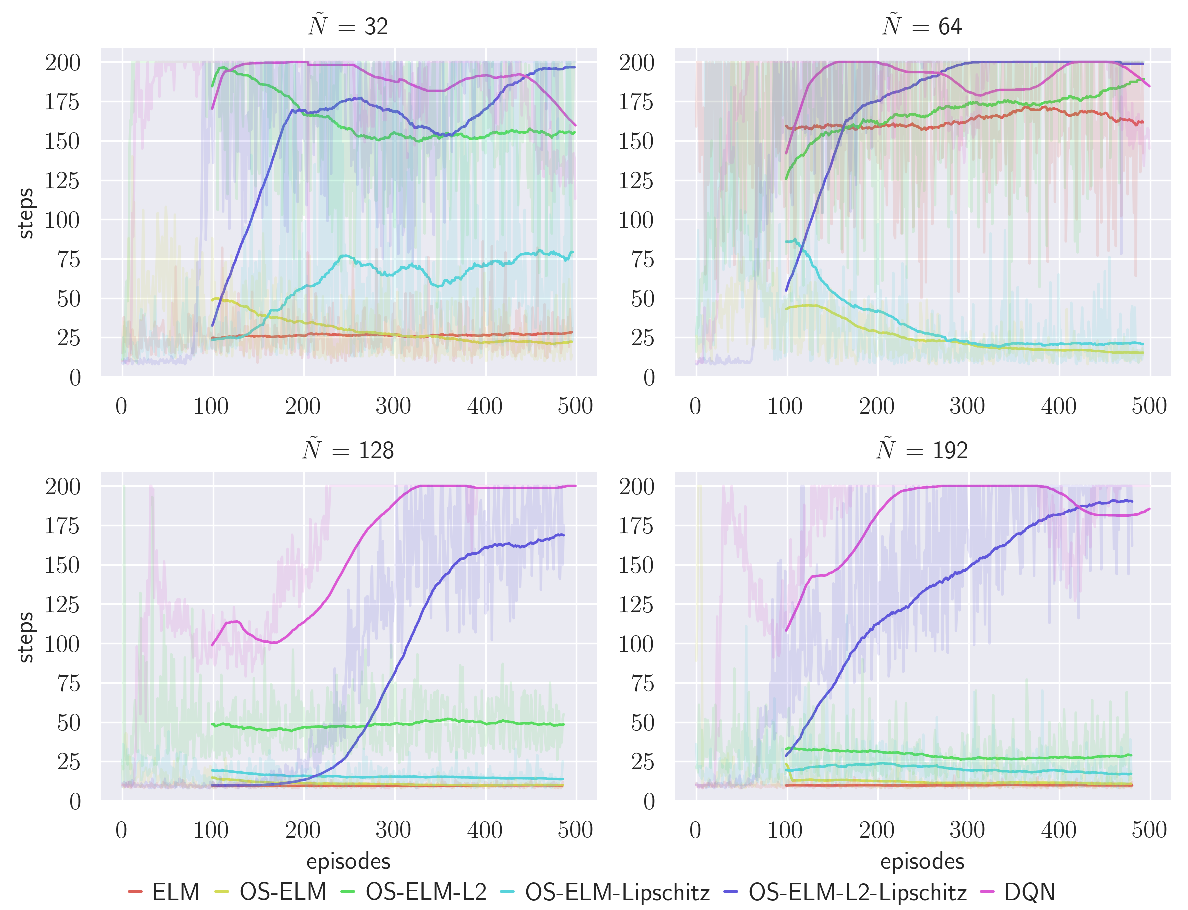}
	\caption{Training curve (light-colored lines: \# of steps
		for continuously standing in each episode; highly-colored
	lines: moving average over last 100 episodes)}
	\label{fig:eval_time}
\end{figure*}

%
%
Figure \ref{fig:eval_time} illustrates training curves of the six
designs when the number of hidden-layer nodes $\tilde{N}$ is varied from 32 to 192.
X-axis shows the number of episodes elapsed, and 
Y-axis shows the number of continuous steps that the inverted pendulum
is standing (higher is better).
There are two line types for each design.
Light-colored lines show the number of steps for continuously standing
in each episode, and highly-colored lines show the moving average over
the last 100 episodes.
In these graphs, a representative result is picked
up for each design for illustration purpose.
Average execution time to complete the task is evaluated in
Section \ref{ssec:eval_time}.
		
%
%
The upper left graph shows the results when the number of hidden-layer
nodes is 32.
In this case, in addition to the baseline {\bf DQN}, the proposed
OS-ELM Q-Networks with regularization and/or normalization techniques
({\bf OS-ELM-L2} and {\bf OS-ELM-L2-Lipschitz}) acquire better actions
that can make the inverted pendulum stand longer.
In the case of {\bf OS-ELM}, on the other hand, as the number of
episodes increases, the number of steps for continuously standing is
getting worse.
This result demonstrates that the Q-value clipping technique is not
sufficient for the stable reinforcement learning and the
regularization and/or normalization techniques are required.
	
%
%
The reinforcement learning is stable in {\bf OS-ELM-L2-Lipschitz} that
uses both the L2 regularization and spectral normalization.
In this case, a generalization capability is improved by the L2
regularization and an output range is limited by the spectral
normalization.
That is, the L2 regularization works directly on weight parameters
$\bm{\beta}$ which are updated at each step.
The spectral normalization affects $\bm{\alpha}$ so that an output
value range of {\bf OS-ELM-L2-Lipschitz} is less than or equal to
$\sigma_{max}(\bm{\beta})$; thus, outliers due to $\bm{\alpha}$ values
can be suppressed by the spectral normalization.
Please note that even if rewards of {\bf OS-ELM-L2-Lipschitz} are
declined once, it can recover the situation and then get right actions.
		
%
%
The upper right graph shows the results when the number of
hidden-layer nodes is 64.
A similar tendency mentioned above is observed in this case too, but
{\bf ELM} also acquires correct actions, because it is expected that
this configuration ($\tilde{N} = 64$) is best suited for {\bf ELM}.

%
%
The lower two graphs show the results when the numbers of hidden-layer
nodes are 128 and 192.
These results are similar.
Only {\bf DQN} and the proposed {\bf OS-ELM-L2-Lipschitz} can acquire
correct actions.
{\bf OS-ELM-L2} and {\bf OS-ELM-Lipschitz} fail to learn correct
actions, indicating that using either the L2 regularization or the
spectral normalization is not sufficient.
In summary, {\bf OS-ELM-L2-Lipschitz} can avoid the overfitting
situation and acquire correct actions thanks to the constraints
on both $\bm{\alpha}$ and $\bm{\beta}$.
		
\subsection{Execution Time to Complete}\label{ssec:eval_time}

\begin{figure*}[t!]
	\centering
	\subfigure[OS-ELM based approaches\label{fig:eval_reward_oselms}]{
		\includegraphics[height=76mm]{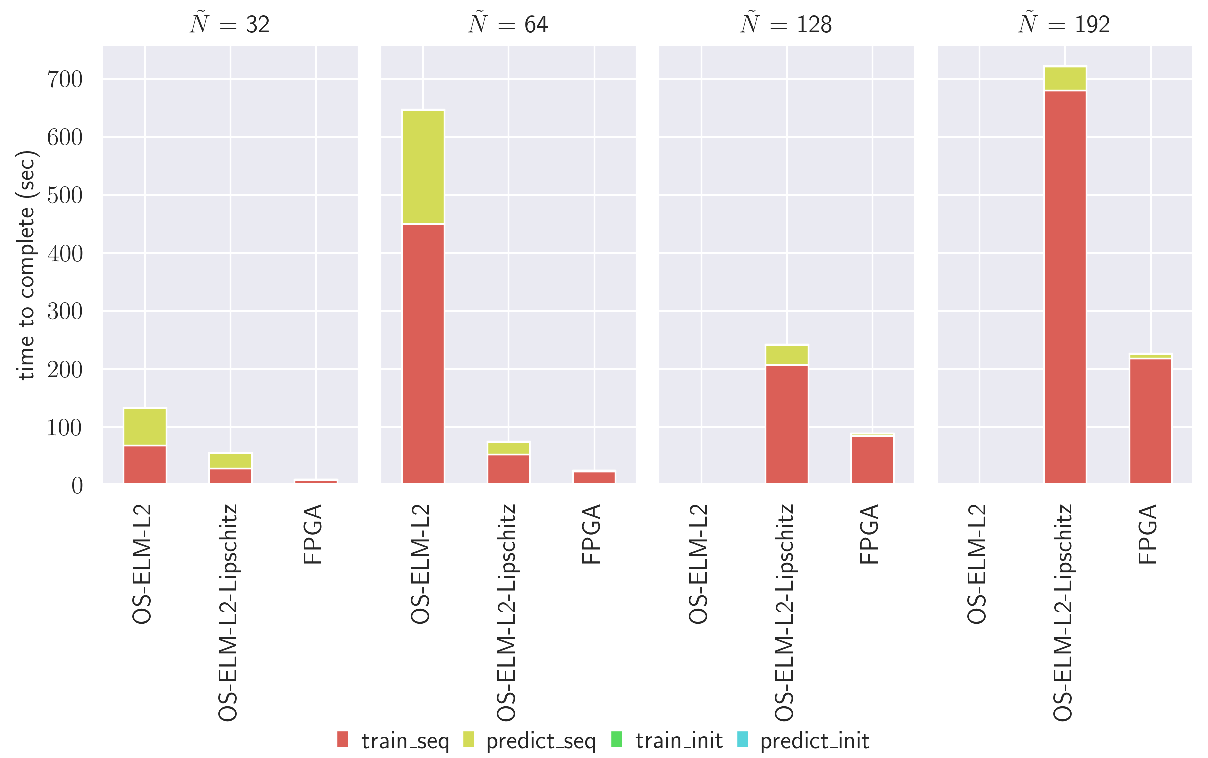}
	}
	\subfigure[{\bf DQN}\label{fig:eval_reward_dqn}]{
		\includegraphics[height=76mm]{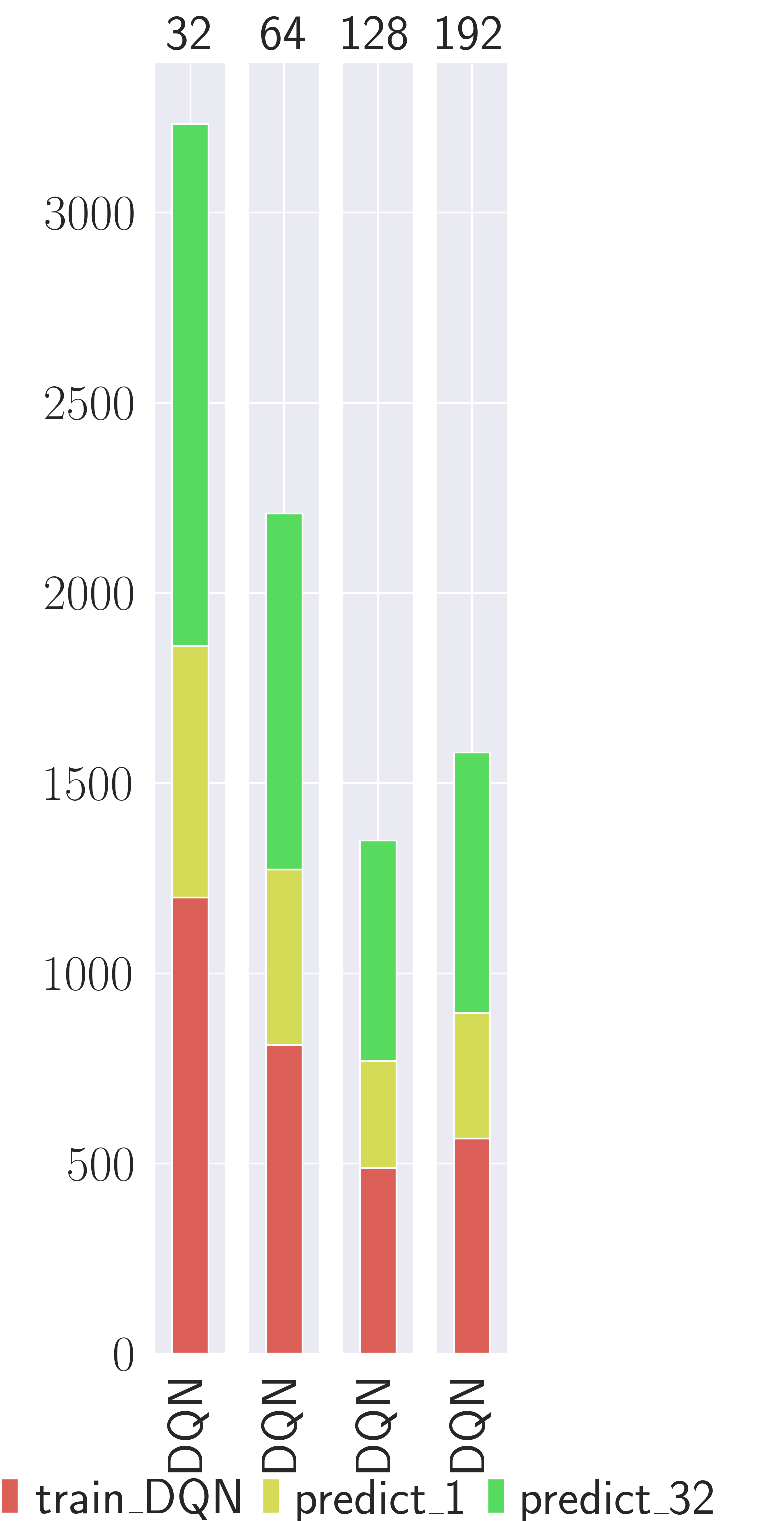}
	}
	\caption{Execution time to complete [sec]}
	\label{fig:eval_reward}
\end{figure*}

%
%
We evaluate the seven designs in terms of execution times to complete
the CartPole-v0 task when the number of hidden-layer nodes $\tilde{N}$
is varied from 32 to 192.
In this evaluation, an execution was terminated as ``impossible'' if
it could not complete the task after 50,000 episodes.
As a result, {\bf OS-ELM} and {\bf OS-ELM-Lipschitz} could not
complete the task in our evaluation.
Also, {\bf ELM} was not stable.
Figure \ref{fig:eval_reward} shows the execution times of 
{\bf OS-ELM-L2}, {\bf OS-ELM-L2-Lipschitz}, {\bf DQN}, and {\bf FPGA}.
{\bf DQN} is separated in the graph since its execution time is quite
large compared to the others.
Table \ref{tb:eval_reward_fixed} shows detailed breakdown of the
proposed {\bf FPGA} design.
	
%
%
In these graphs, each bar shows the execution time breakdown of each
operation: train\_seq, predict\_seq, train\_init, predict\_init,
train\_DQN, predict\_1, and predict\_32.
\begin{itemize}
	\item In the OS-ELM based approaches except for {\bf FPGA},
	      train\_init and train\_seq indicate their initial training and
	      sequential training, respectively.
	      predict\_init and predict\_seq are their predictions before and
	      after their initial training is completed, respectively.
	      All these operations are done by the CPU part.
	\item In the proposed {\bf FPGA}, before the initial training,
	      train\_init and predict\_init (Steps 2a and 2c) are executed by the
	      CPU part.
	      After the initial training, train\_seq and predict\_seq (Steps 4a
	      and 4c) are done by the FPGA part.
PS and PL parts are connected via AXI bus and DMA transfer is
used for their communication though not fully implemented in our design.
	      We assume data transfer latency between the CPU and FPGA parts
	      is 1 cycle per float32.
	      This is an optimistic assumption, but we use this value for
	      simplicity because it varies depending on an underlying
	      hardware platform (e.g., DMA performance).
	\item In the baseline {\bf DQN}, train\_DQN is its training.
	      predict\_1 and predict\_32 indicate its predictions when the batch
	      sizes are 1 and 32, respectively.
	      More specifically, predict\_1 and predict\_32 are called from
	      Determine and Update states, respectively.
	      All the operations are done by the CPU part.
\end{itemize}
Execution time for interactions with a given environment (Steps 2b and
4b) is not considered in this evaluation.
train\_init and predict\_init exist but are negligible.


\begin{table}[t!]
	\centering
	\caption{Execution time to complete (breakdown of {\bf FPGA}) [sec]}
	\begin{tabular}{c|rrrr|r}\hline \hline
		$\tilde{N}$ & train\_seq & predict\_seq & train\_init & predict\_init & Total   \\ \hline
		32          & 7.847      & 1.466        & 0.023       & 0.053         & 9.389   \\
		64          & 22.458     & 2.135        & 0.047       & 0.067         & 24.707  \\
		128         & 84.038     & 4.036        & 0.245       & 0.166         & 88.484  \\
		192         & 218.258    & 7.005        & 0.685       & 0.281         & 226.230 \\\hline
	\end{tabular}
	\label{tb:eval_reward_fixed}
\end{table}
	
%
%
The breakdown of each operation is computed by (the number of
executions of the operation) $\times$ (execution time of the single
operation).
train\_seq is dominant compared to train\_init because 
train\_init is executed only once for each episode.
These execution times are averaged over 150 trials. 
	
%
%
When the number of hidden-layer nodes is 32, {\bf OS-ELM-L2}, 
{\bf OS-ELM-L2-Lipschitz}, {\bf DQN}, and {\bf FPGA} can acquire
correct actions.
Their execution times are 132.27sec, 55.02sec, 3232.54sec, 
and 9.39sec, respectively.
%
%
When the number of hidden-layer nodes is 64, {\bf OS-ELM-L2}, 
{\bf OS-ELM-L2-Lipschitz}, {\bf DQN}, and {\bf FPGA} can acquire 
correct actions.
Their executions times are 647.56sec, 74.20sec, 2208.90sec, 
and 24.71sec, respectively.
The execution times of {\bf OS-ELM-L2}, {\bf OS-ELM-L2-Lipschitz}, and
{\bf FPGA} are increased compared to their previous result having 32
hidden-layer nodes because of a larger matrix size.
In this case, {\bf OS-ELM-L2}, {\bf OS-ELM-L2-Lipschitz}, and the
proposed {\bf FPGA} are faster than {\bf DQN} by 3.41x, 29.77x, and
89.40x, respectively.
		
%
%
		
%
%
		
%
%
These results demonstrate that
{\bf FPGA} is the fastest followed by 
{\bf OS-ELM-L2-Lipschitz} and {\bf DQN}, because update formula of
the OS-ELM based approaches is simple as shown in Equations
\ref{eq:os_elm_0} and \ref{eq:r_oselm_1}.
Although {\bf FPGA} and {\bf OS-ELM-L2-Lipschitz} use the same
algorithm, {\bf FPGA} is faster, because train\_seq and
predict\_seq are accelerated by dedicated circuits, as shown in Figure
\ref{fig:design_overview}.
%
%
Regarding the performance bottleneck, the OS-ELM based approaches
spend most of time for train\_seq, while {\bf DQN} spends a certain
time for train\_DQN, predict\_1, and predict\_32.
%
%
As mentioned above, the execution times tend to increase as the
number of hidden-layer nodes is increased except for {\bf DQN}.
This is because the size of matrix products is denoted as
$\mathbb{R}^{\tilde{N}\times \tilde{N}}\cdot\mathbb{R}^{\tilde{N}\times \tilde{N}}$,
and the computation cost increases rapidly as the number of
hidden-layer nodes is increased.
%
%
Such matrix products can be accelerated efficiently by dedicated
logic; thus, the proposed {\bf FPGA} design is advantageous
for the on-device reinforcement learning on resource-limited
edge devices.
	

\section{Summary}\label{sec:conc}

%
%
To solve reinforcement learning tasks on resource-limited edge
devices, in this paper, we proposed OS-ELM Q-Network as a lightweight
reinforcement learning algorithm that do not rely on a backpropagation
based iterative optimization.
More specifically, the following techniques were proposed for OS-ELM
Q-Network:
(1) simplified output model, (2) Q-value clipping, (3) random update,
and (4) combination of the spectral normalization for $\bm{\alpha}$
and L2 regularization for $\bm{\beta}$.
Especially, thanks to (4), the Lipschitz constant of OS-ELM can be
suppressed under $\sigma_{max}(\bm{\beta})$ and further controlled by
adjusting the parameter $\delta$.

%
%
OS-ELM Q-Network with all the above techniques was designed for
PYNQ-Z1 board as a low-cost FPGA platform by extending an existing
on-device learning core \cite{tukada_oselm}.
Prediction and sequential training in most of Determine and Update
states (i.e., predict\_seq and train\_seq) are accelerated by the FPGA
part, and the others are executed by the CPU part.
%
%
The evaluation results using OpenAI Gym demonstrated that the proposed
{\bf OS-ELM-L2-Lipschitz} and its FPGA implementation
complete a CartPole-v0 task 29.77x and 89.40x faster than a
conventional DQN-based approach when the number of hidden-layer nodes
is 64.
Also, they are robust against the number of hidden-layer nodes thanks
to (4).

%
%




\begin{thebibliography}{10}

\bibitem{dqn}
V.~Mnih, K.~Kavukcuoglu, D.~Silver, A.~Graves, I.~Antonoglou, D.~Wierstra, and
  M.~Riedmiller, ``{Playing Atari with Deep Reinforcement Learning},''
  \emph{arXiv:1312.5602}, Dec 2013.

\bibitem{experience_replay}
L.-J. Lin, ``{Reinforcement Learning for Robots Using Neural Networks},'' Ph.D.
  dissertation, Carnegie Mellon University, USA, Jan 1993.

\bibitem{tukada_oselm}
M.~Tsukada, M.~Kondo, and H.~Matsutani, ``{A Neural Network-Based On-device
  Learning Anomaly Detector for Edge Devices},'' \emph{IEEE Transactions on
  Computers}, vol.~69, no.~7, pp. 1027--1044, Jul 2020.

\bibitem{os_elm}
N.-Y. Liang, G.-B. Huang, P.~Saratchandran, and N.~Sundararajan, ``{A Fast and
  Accurate Online Sequential Learning Algorithm for Feedforward Networks},''
  \emph{{IEEE Transactions on Neural Networks}}, vol.~17, no.~6, pp.
  {1411--1423}, Nov 2006.

\bibitem{universal_approximation_theorem}
K.~Hornik, M.~Stinchcombe, and H.~White, ``{Multilayer Feedforward Networks are
  Universal Approximators},'' \emph{Neural Networks}, vol.~2, no.~5, pp. 359 --
  366, Jul 1989.

\bibitem{spectral_normalization}
T.~Miyato, T.~Kataoka, M.~Koyama, and Y.~Yoshida, ``{Spectral Normalization for
  Generative Adversarial Networks},'' in \emph{Proceedings of the International
  Conference on Learning Representations (ICLR'18)}, Feb 2018.

\bibitem{elm}
G.-B. Huang, Q.-Y. Zhu, and C.-K. Siew, ``{Extreme Learning Machine: A New
  Learning Scheme of Feedforward Neural Networks},'' in \emph{{Proceedings of
  the International Joint Conference on Neural Networks (IJCNN'04)}}, Jul 2004,
  pp. {985--990}.

\bibitem{r_oselm}
H.~T. Huynh and Y.~Won, ``{Regularized Online Sequential Learning Algorithm for
  Single-Hidden Layer Feedforward Neural Networks},'' \emph{Pattern Recognition
  Letters}, vol.~32, no.~14, pp. 1930 -- 1935, Oct 2011.

\bibitem{dqn_nature}
V.~Mnih \emph{et~al.}, ``{Human-Level Control through Deep Reinforcement
  Learning},'' \emph{Nature}, vol. 518, no. 7540, pp. 529--533, Feb 2015.

\bibitem{spinningup}
J.~Achiam, ``{Spinning Up in Deep Reinforcement Learning},''
  \url{https://github.com/openai/spinningup}, 2018.

\bibitem{spectral_regularization}
Y.~Yoshida and T.~Miyato, ``{Spectral Norm Regularization for Improving the
  Generalizability of Deep Learning},'' \emph{arXiv:1705.10941}, May 2017.

\bibitem{adam}
D.~P. Kingma and J.~Ba, ``{Adam: A Method for Stochastic Optimization},'' in
  \emph{Proceedings of the International Conference on Learning Representations
  (ICLR'15)}, May 2015.

\bibitem{huber}
P.~J. Huber, ``{Robust Estimation of a Location Parameter},'' \emph{Annals of
  Mathematical Statistics}, vol.~35, no.~1, pp. 73--101, Mar 1964.

\end{thebibliography}

\end{document}